\pdfoutput=1

\documentclass[11pt]{article}

\usepackage{acl}

\usepackage{times}
\usepackage{latexsym}
\usepackage{graphicx}
\usepackage{makecell}
\usepackage{caption}
\usepackage{subcaption}
\usepackage[many]{tcolorbox}  
\usepackage{multirow}

\usepackage[T1]{fontenc}

\usepackage[utf8]{inputenc}

\usepackage{microtype}
\usepackage{cleveref}
\usepackage{array}

\newcolumntype{P}[1]{>{\centering\arraybackslash}p{#1}}
\newcolumntype{M}[1]{>{\centering\arraybackslash}m{#1}}

\newtcolorbox{boxA}{
    fontupper = \small,
    fontlower = \small,
    boxrule = 0.5pt,
    colframe = black, 
    width=7.5cm,
    text width=7cm,
    parbox=false,
    halign=left,
}

\title{Explainable Verbal Reasoner Plus (EVR+): A Natural Language Reasoning Framework that Supports Diverse Compositional Reasoning}

\author{Zhengzhong Liang$^{\spadesuit}$, \, Zeyu Zhang$^{\clubsuit}$, \, Steven Bethard$^{\clubsuit}$ \and Mihai Surdeanu$^{\spadesuit}$  \\
  $^{\spadesuit}$Computer Science Department, University of Arizona, Tucson, AZ, USA \\
  $^{\clubsuit}$School of Information, University of Arizona, Tucson, AZ, USA \\
  \texttt{$\{$zhengzhongliang, zeyuzhang, bethard, msurdeanu$\}$@arizona.edu} \\}

\begin{document}
\maketitle
\begin{abstract}
 Languages models have been successfully applied to a variety of reasoning tasks in NLP, yet the language models still suffer from compositional generalization. In this paper we present Explainable Verbal Reasoner Plus (EVR+), a reasoning framework that enhances language models' compositional reasoning ability by (1) allowing the model to explicitly generate and execute symbolic operators, and (2) allowing the model to decompose a complex task into several simpler ones in a flexible manner. Compared with its predecessor Explainable Verbal Reasoner (EVR) \cite{liang2021explainable} and other previous approaches adopting similar ideas, our framework supports more diverse types of reasoning such as nested loops and different types of recursion. To evaluate our reasoning framework, we build a synthetic dataset with five tasks that require compositional reasoning. Results show that our reasoning framework can enhance the language model's compositional generalization performance on the five tasks, using a fine-tuned language model. We also discussed the possibility and the challenges to combine our reasoning framework with a few-shot prompted language model.\footnote{The code is available at: \hfill \\ \url{https://github.com/zhengzhongliang/ExplainableVerbalReasonerPlus.git}}
\end{abstract}

\section{Introduction}

\begin{figure}[!ht]
    \centering
    \includegraphics[width=0.35\textwidth]{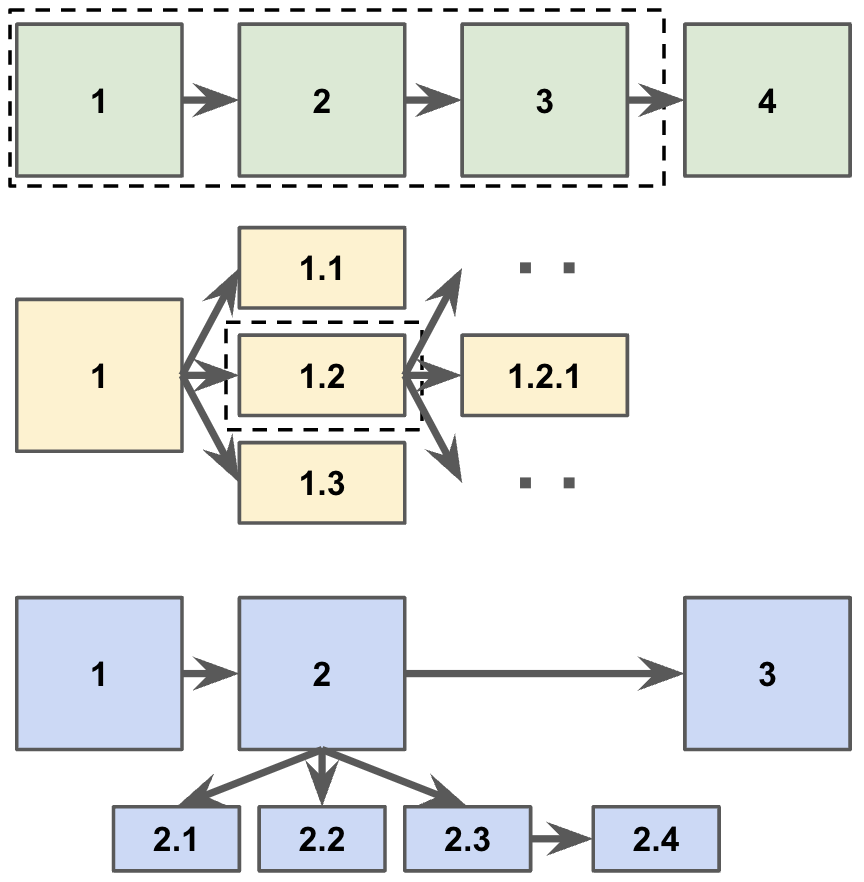}
    \caption{Comparison of Different Iterative Decomposition Frameworks. Top: sequential decomposition. The next action/sub-problem is determined by all previous actions/results. For example, block 4 is generated based on blocks [1,2,3]. Middle: recursive decomposition; the generated action/sub-problem is only determined by the upper level's state (e.g., 1.2.1 is only determined by 1.2); Bottom: hybrid decomposition, in which the hybrid reasoning framework supports both sequential decomposition and recursive decomposition. [2.1, 2.2, 2.3] are generated from [2] recursively; [3] is generated from [1, 2] sequentially.}
    \label{fig:decomposition_comparison}
\end{figure}

Significant progress has been made in applying language models for reasoning tasks in Natural Language Processing, including arithmetic reasoning \cite{cobbe2021training, weichain, wang2022self}, logical reasoning \cite{clark2021transformers, liang2021explainable, tafjord2020proofwriter, richardson2022pushing, creswell2022selection} and reading comprehension problems that require multi-hop reasoning \cite{khot2020text, dua2022successive, khot2022decomposed}. Yet the language models still suffer from the compositional generalization problem: when the language model is trained on simpler tasks, the model tends to fail when tested on harder tasks that are composed of several simpler tasks \cite{clark2021transformers, liang2021explainable, ontanon2022making, weichain}. 

Two efforts have been made to improve the compositional generalization performance of language models. First, symbolic operators are introduced to explicitly guide the language models to perform symbolic operations, such as loops and list operations \cite{wolfson2020break, khot2022decomposed}. Second, language models are taught to decompose a complex problem into several simpler problems. Each decomposed problem can be solved individually by calling a pre-defined module (which also might involve a language model) \cite{khot2020text, liang2021explainable, khot2022decomposed}. Empirical experiments show that these two designs can effectively improve language models' compositional generalization performance on a variety of tasks. However, there are two drawbacks of the current approaches. First, the symbolic operators are usually defined for specific tasks and it is hard to apply the operators to a wide range of tasks. For example, the \texttt{for$\_$each} operator used in \cite{khot2022decomposed} can loop over a list, but it does not naturally support more complex loops such as nested loops or loops with inner if-else conditional statements. Second, the current decomposition methods do not allow the decomposed problems to be further decomposed into a few subproblems in a flexible enough manner. For example, \cite{dua2022successive} decomposes a complex question answering question into a few subquestions, where each sub question can be directly answered by a neural QA model. However, this method does not support the further decomposition of the subquestions.

In this work, we introduce Explainable Verbal Reasoner Plus (EVR+), a reasoning framework that can be applied to fairly diverse compositional reasoning tasks. EVR+ works in cycles: in each cycle, EVR+ first generates a short program, then the program is parsed and executed. A complex problem is solved after a few such cycles. This is similar to the design of its predecessor, Explainable Verbal Reasoner (EVR). However, the programs that can be generated and executed by EVR are limited and task-specific; therefore, EVR could only be applied to solve backward logic reasoning with tree search. In contrast, EVR+ can generate and parse much more complex programs including complex statements such as nested loops, conditions, list operations and diverse types of recursion. Therefore EVR+ can be used to handle much broader types of reasoning. 


We evaluate EVR+ on a newly introduced synthetic dataset, SynthCompR, which contains five tasks that requires compositional reasoning: chaining, Cartesian product, tree search, chaining tree search and Cartesian tree search (examples in Table \ref{tab:dataset_examples}). These tasks aim at testing the model's ability to handle different types of compositional reasoning, such as loops and recursion. Results show that applying EVR+ results in considerably better generalization performance on the reasoning tasks than using a standard UnifiedQA-T5-large model.

In summary, our contributions are: (1) we design five diagnostic synthetic tasks to evaluate the models' performance on compositional reasoning; (2) we present a reasoning framework which supports a diverse types of compositional reasoning; (3) we show that applying our reasoning framework yields better compositional generalization performance on the five reasoning tasks compared to and end-to-end trained language model.

\begin{table*}
    \centering
    \begin{tabular}{ c | c | c | c | c | c | c } \hline
         & \multicolumn{3}{c |}{du2/du3} & \multicolumn{3}{c}{du4} \\ \hline
         & train & dev & test & train & dev & test \\ \hline \hline
        chaining & 9999 & 999 & 999 & 10000 & 1000 & 1000 \\
        Cartesian & 10000 & 1000 & 1000 & 9999 & 999 & 999 \\
        tree search & 9999 & 999 & 999 & 10000 & 1000 & 1000 \\
        chaining tree search & 9999 & 999 & 999 & 10000 & 1000 & 1000 \\ 
        Cartesian tree & 9999 & 999 & 999 & 10000 & 1000 & 1000 \\ \hline
    \end{tabular}
    \caption{Statistics of the SynthCompR Dataset. For the tasks except the Cartesian task we generate the du2 and du4 data. Du means depth up to (e.g., du2 means depths 0, 1, 2). For the Cartesian product task, we generate the du3 and du4 data. The number of examples of each depth are equally distributed. For example, the chaining du2 training split has 9999 examples, each depth [0, 1, 2] has 3333 training examples. }
    \label{tab:dataset_statistics}
\end{table*}

\section{Related Work}
\paragraph{Compositional Generalization}
Previous studies show that neural networks tend to suffer from compositional generalization: when the neural models are trained to parse simpler expressions, they have difficulties generalize to harder expressions that are composed using the same rules as the simpler expressions \cite{lake2018generalization, hupkes2020compositionality, ontanon2022making}. Similar findings are also reported for logic reasoning problems \cite{sinha2019clutrr, clark2021transformers, liang2021explainable, tafjord2020proofwriter}. One remedy is to design neural networks with task-specific architectures that have the inductive bias suited for the specific task. However, previous study shows that it is hard to design end-to-end trained neural networks that can achieve satisfactory compositional generalization on a wide range of tasks \cite{furrer2020compositional, ontanon2022making}. Another direction is to teach the neural networks to explicitly generate the symbolic operators which can be later processed and executed. However, the existing symbolic operators are usually designed for a specific task and not suitable for diverse tasks \cite{guo2019towards, furrer2020compositional, liang2021explainable}. In contrast, our reasoning framework allows the neural network to generate the operators that can be applied to diverse types of tasks.

\paragraph{Problem Decomposition} Iteratively decomposing a complex problem into several simpler problems is actively being investigated by many researchers. There are mainly two types of methods: sequential decomposition and recursive decomposition (Figure \ref{fig:decomposition_comparison}). In sequential decomposition, the next decomposition to perform is determined by the previous decompositions (and optionally the results to all previous decompositions) \cite{khot2020text, yang2022seqzero, dua2022successive, wang2022iteratively}. On the other hand, the recursive decomposition approach generates multiple decompositions at once, and the generated decompositions are only dependent on the upper level's state (not all previous steps) \cite{
kazemi2022lambada, liang2021explainable}. Our reasoning framework supports both sequential decomposition and recursive decomposition in a very flexible manner (i.e., hybrid decomposition). A detailed comparison between our reasoning framework and the most similar framework is given in Section \ref{sec:discussion}.

\section{SynthCompR: A Synthetic Dataset for Compositional Reasoning}
We first present a synthetic dataset that can be used to evaluate the models' ability of compositional reasoning. To include diverse types of reasoning problems, we include five tasks in the dataset: \textbf{chaining}, \textbf{Cartesian}, \textbf{tree search}, \textbf{chaining tree search}, \textbf{Cartesian tree search}. All of these tasks are motivated by real-world examples or previously proposed problems that require multiple steps of reasoning. We define the steps of reasoning as \textbf{depth}. We construct the dataset so that each task consists of problems with different depths. The task is in the question answering format, where each problem has a context, a question and an answer. Table \ref{tab:dataset_statistics} shows the statistics of the dataset. Examples of these tasks are shown in Table \ref{tab:dataset_examples} and detailed below.\footnote{The description of the dataset construction process is given in Appenidx \ref{sec:app_dataset_construction}.}

\begin{table*}[!ht]
    \small
    \centering
    \begin{tabular}{M{1cm} | M{8.5cm} | M{2cm} | M{3.5cm}} \hline
        Task & Context & Question & Answer \\ \hline
         Chaining & [\textbf{Chains}] Kevin Cox had 15 apples in the beginning. Terry Rivera had 15 apples in the beginning. Robert Murphy had 18 apples in the beginning. Eugene Gray gave Kevin Cox 1 apple. Arthur Clark gave Kevin Cox 1 apple. Chris Wilson did not give Terry Rivera any apples. Chris Edwards did not give Terry Rivera any apples. Shawn Torres gave Robert Murphy 2 apples. Robert Murphy gave Gerald Bennett 1 apple. & How many apples did Robert Murphy have in the end? & 19 \\ \hline 
        Cartesian & [\textbf{Fact}] Each of Matthew Rodriguez and Jonathan Diaz had 8 bananas and 15 puppies. & List the items that each person had. & Matthew Rodriguez had 8 bananas, Matthew Rodriguez had 15 puppies, Jonathan Diaz had 8 bananas, Jonathan Diaz had 15 puppies. \\ \hline 
        Tree search & [\textbf{Facts}] Ralph James had 18 peaches. Gary White had 18 peaches. Gary White had 19 apples. [\textbf{Rules}]  If Ralph James had 18 pears then Matthew Smith had 7 rabbits. If Ralph James had 18 pears then Kenneth Reed had 14 bananas. If Ralph James had 12 peaches then Dennis Watson had 6 kittens. If Gary White had 18 peaches then Matthew Smith had 8 bananas. If Gary White had 19 apples then Kenneth Reed had 14 bananas. & Did Matthew Smith have 8 bananas? & Yes \\ \hline 
        Chaining tree search & [\textbf{Chains}] Sean Sanchez had 3 puppies in the beginning. Sean Sanchez had 3 owls in the beginning. Andrew Flores had 4 owls in the beginning. Nicholas Gray had 14 puppies in the beginning. Sean Sanchez gave Antonio Evans 2 puppies. Billy Ross did not give Sean Sanchez any owls. Joshua Carter gave Andrew Flores 2 owls. Kenneth Howard did not give Nicholas Gray any puppies. [\textbf{Rules}] If Andrew Flores had 1 owl then Jeremy Morgan had 4 puppies. If Sean Sanchez had 18 owls then Bobby Garcia had 14 puppies. If Sean Sanchez had 1 puppy then Jeremy Morgan had 15 puppies. If Nicholas Gray had 14 puppies then Jeremy Morgan had 8 kittens. If Sean Sanchez had 18 owls and Nicholas Gray had 14 puppies then Jeremy Morgan had 4 puppies. & Did Jeremy Morgan have 15 puppies? & Yes \\ \hline 
        Cartesian tree search & [\textbf{Fact}] Each of Johnny Gomez, David Rogers and Howard Morgan had 8 pens, 13 peaches and 14 kittens. [\textbf{Rules}] If Craig Ward had 13 peaches and Aaron Hughes had 2 pens then George Smith had 14 toy cars. If Howard Morgan had 14 kittens then Kenneth King had 14 owls. If Craig Ward had 13 peaches and Aaron Hughes had 2 pens then Keith Rogers had 14 owls. If Aaron Hughes had 2 pens then Joshua Morris had 14 owls. If Johnny Gomez had 14 kittens then Keith Rogers had 14 owls. & Did Joshua Morris have 14 owls? & No \\ \hline 
    \end{tabular}
    \caption{The 5 synthetic tasks that are motivated by real-world examples or previously proposed problems that require compositional reasoning. }
    \label{tab:dataset_examples}
\end{table*}

\subsection{Chaining}
Testing the model's ability to keep track of the state of objects in the world? has been explored in multiple previous works \cite{weston2015towards, dalvi-etal-2019-everything, cobbe2021training, weichain, tamari2022dyna}. A simple question from the ProPara dataset \cite{dalvi-etal-2019-everything} is ``\textit{Trash is put into a household trashbag. The trashbags are thrown into a large outdoor trashcan. The trashcan is emptied by a large trash truck. The trash truck travels to the landfill and unloads all the trash the truck has collected. Where is the trash in the end?}''. To answer this question, the model needs to (explicitly or implicitly) construct the chain of states of ``\textit{trash}'': ``\textit{trashbag $\rightarrow$ trashcan $\rightarrow$ trash truck $\rightarrow$ landfill}''. 

We similarly include a chaining task in our dataset, where an object's initial state is updated multiple times given a chain of statements and the model needs to infer the final state of the object. Table \ref{tab:dataset_examples} shows a chaining problem with depth 2, i.e., ``\textit{the number of apples Robert Murphy had}'' is updated two times: \textit{Robert Murphy had 18 apples in the beginning $\rightarrow$ Shawn Torres gave Robert Murphy 2 apples $\rightarrow$ Robert Murphy gave Gerald Bennett 1 apple.}

\subsection{Cartesian Product}
The ability to perform Cartesian Product is an essential part of various reasoning tasks. For example, consider a semantic parsing problem of the query ``who directed and acted in M1 and M2?'', the model needs to convert the query to four queries: ``directed M1'', ``directed M2'', ``acted in M1'' and ``acted in M2'' \cite{keysers2019measuring, ontanon2022making}. We similarly include a Cartesian product task in our dataset. Table \ref{tab:dataset_examples} shows an example with depth 2 (because there are two persons and two types of items).

\subsection{Tree Search}
The ability to perform tree search is essential to planning and reasoning. The most ordinary case for tree search in the NLP community is logical reasoning, where forward chaining and backward chaining are used to prove a statement \cite{dalvi2021explaining, ribeiro2022entailment, bostrom2022natural, hong2022metgen, liang2021explainable, tafjord2020proofwriter, qu2022interpretable, creswell2022selection, sanyal2022fairr, creswell2022selection}. Table \ref{tab:dataset_examples} shows an tree search example with depth 1, because the statement ``\textit{Matthew Smith had 8 bananas}'' can be proved using one rule: ``\textit{If Gary White had 18 peaches then Matthew Smith had 8 bananas}''. And then ``\textit{Gary White had 18 peaches}'' is verified by the given facts. A tree search problem with a large depth means multiple rules need to be applied iteratively to prove a statement. When building the dataset, we explicitly build some examples that require backtracking. For example, both rules \textit{if B then A} and \textit{if C then A} can potentially be used to prove a statement \textit{A}. However, if statement \textit{B} cannot be proved by other rules and facts, the model needs to backtrack and continue to try to verify statement \textit{C}.

\subsection{Chaining Tree Search}
A natural rising question of compositional reasoning is: if a model has mastered the skills to solve some simple problems, can the model solve some hard problems that are composed of the simple problems? By combining the chaining task and the tree search task, we include the chaining tree search task, where both chaining and tree search are needed to answer the question. Table \ref{tab:dataset_examples} shows a chaining tree search problem with depth 1. To answer the question ``\textit{Did Jeremy Morgan have 15 puppies?}'', one first needs infer ``\textit{How many items each person had after exchanging the items}'' using the \textbf{Chains} statements. For example, given the facts that ``\textit{Sean Sanchez had 3 puppies in the beginning}'' and ``\textit{Sean Sanchez gave Antonio Evans 2 puppies}'', we can infer ``\textit{Sean Sanchez had 1 puppy}''. And given the \textbf{Rule} ``\textit{If Sean Sanchez had 1 puppy then Jeremy Morgan had 15 puppies}'', we infer that the answer is ``\textit{Yes}''.
\subsection{Cartesian Tree Search}
Similar to the chaining tree search task, we combine the Cartesian problem with the tree search problem to build the Cartesian tree search problem. Table \ref{tab:dataset_examples} shows a Cartesian tree search problem with depth 1. To answer the question ``\textit{Did Joshua Morris have 14 owls?}'', one needs to first infer the number of items each person had given the provided \textbf{Fact}, then use the \textbf{Rules} and the inferred facts to obtain the final answer.

\begin{figure}
    \centering
    \includegraphics[width=0.45\textwidth]{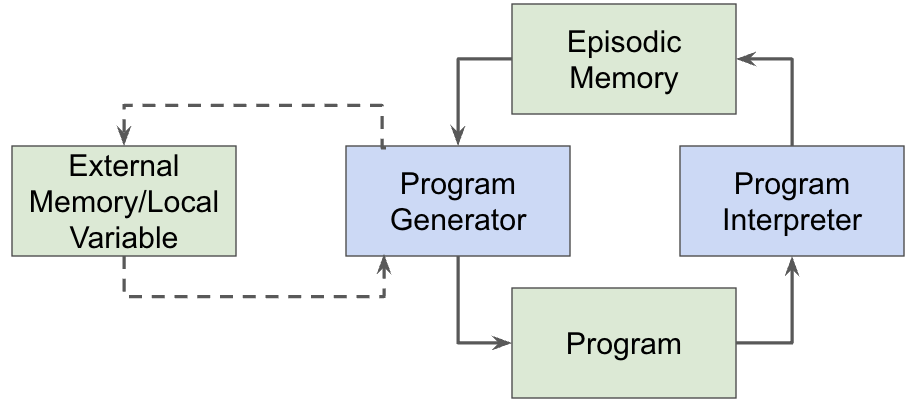}
    \caption{An EVR+ Working Cycle}
    \label{fig:evr_cycle}
\end{figure}

\section{The EVR+ Reasoning Framework}
We propose Explainable Verbal Reasoner Plus (EVR+), a reasoning framework featuring (1) the ability to generate and execute symbolic operators; (2) the ability to decompose a hard problem to a few simpler ones; (3) the ability to support diverse types of compositional reasoning, such as the reasoning covered in SynthCompR and (4) the ability to process natural language input. EVR+ is motivated by the Explainable Verbal Reasoner framework introduced in \cite{liang2021explainable}. EVR works on natural language, and is able to decompose a complex reasoning problem recursively into simpler problems. However, EVR does not support reasoning problems other than recursion, such as chaining or Cartesian product. Extended from EVR, EVR+ is able to support more types of reasoning problems. We achieved this by allowing EVR+ to generate much more diverse and complex operators than EVR, and developed an interpreter that can parse and execute such operations. We detail these contributions next.

\subsection{Framework Overview}
\label{sec:framework_overview}
Figure \ref{fig:evr_cycle} shows a working cycle of EVR+: given the content of the \textbf{episodic memory} (i.e., some textual input), the \textbf{program generator} generates the program, which will be later parsed and executed by the \textbf{program interpreter}. The execution of the program will result in the modification of the episodic memory and the \textbf{local variable buffer}, and less frequently, the modification of the \textbf{external memory}. Then EVR+ enters a new working cycle (generating and executing program). Below are high-level descriptions of the components of EVR+. A detailed walk-through example containing how these components interact with each other is presented in Section \ref{sec:walkthrough}.

\paragraph{Memory}
Three types of memory are used for EVR+: episodic memory, external memory and local variable buffer. Episodic memory stores the result of the executed programs, and the program generator uses the content of the episodic memory as the input to generate program. External memory stores the content that is not likely to be changed frequently. For example, in a reading comprehension task, the external memory can be used to store the raw passages. The local variable buffer stores the temporary variables during the execution of the generated program. But the local variables will not be used by the program generator to generate the program.\footnote{Exchanging the values of two variables can be a good example to show the function of a local variable. Assume $a=1$ and $b=2$, a natural way to swap their values is to set $c=a$, then let $a=b$ and $b=c$. Here $c$ is only used to temporarily store a value. We use local variables in the same way.}

\paragraph{Program Generator}
We primarily used UnifiedQA-T5-base \cite{khashabi2020unifiedqa} as the program generator. The input to the program generator is the textual statements in the episodic memory. The program generator then generates program based on the textual input, then the program will be handled by the interpreter. However, the EVR+ framework is largely model-agnostic. Theoretically other generative language models such as BART \cite{lewis2019bart} can also be used as the program generator. Section \ref{sec:llm_compatibility} also discusses the possibility and the challenges of combining EVR+ with a few-shot prompted GPT-3 \cite{brown2020language}.

\paragraph{Program Interpreter}
The program interpreter parses the generated program and executes it. One advancement of EVR+ compared to EVR is that EVR+ supports much more diverse expressions than EVR.\footnote{A discussion about the relation between our interpreter and python is discussed in Section \ref{sec:relation_with_python}} Table \ref{tab:list_of_expressions_1} and \ref{tab:list_of_expressions_2} in Appendix \ref{sec:appendix_operations} show the complete list of supported operations that can be parsed by our interpreter. In general, the following types of expressions are supported by EVR+.

\begin{itemize}
    \item \textbf{Constants} such as numbers, strings and Boolean values.
    \item \textbf{Local variables} that are used to store the temporary values during the execution of the program.
    \item \textbf{Control flows} such as \texttt{for} loop, \texttt{while} loop and \texttt{if-else} conditions.
    \item \textbf{Memory operations} such as the read and write operations to the episodic memory, local variable buffer, external memory and the \texttt{new$\_$mem()} operator that can start a new recursion process.
    \item \textbf{Tools}, which are modules designed to fulfill a specific functionality, such as \texttt{qa} and \texttt{rewrite}. 
\end{itemize}
The example usages of these operations can be found in Figure \ref{fig:walkthrough_tree_search}.

\subsection{Tree Search: A Walk-through Example}
\label{sec:walkthrough}
\begin{figure*}[!ht]
    \centering
    \includegraphics[width=0.98\textwidth]{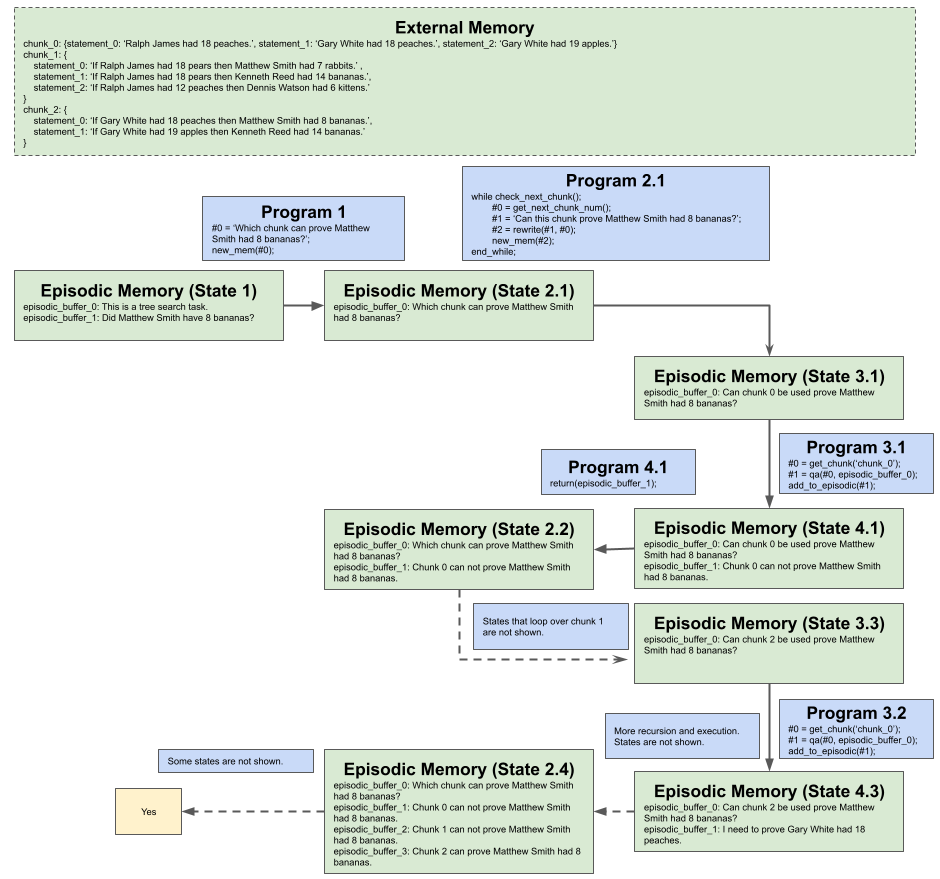}
    \caption{A walk-through example of EVR+ on the tree search task. }
    \label{fig:walkthrough_tree_search}
\end{figure*}

We show a walk-through example for the tree search problem using EVR+ in this section (Figure \ref{fig:walkthrough_tree_search}). Appendix \ref{sec:app_execution_flows} shows additional walk-through examples for all tasks. Note that the execution flow in Figure \ref{fig:walkthrough_tree_search} involves neural components, and the shown execution flow assumes the neural components have been trained. The training process of the neural components is described in Section \ref{sec:framework_training}.

In Figure \ref{fig:walkthrough_tree_search}, every two states connected by a solid line represent a \textbf{working cycle} as defined in Section \ref{sec:framework_overview}. For example, the transition from \textbf{State 1} to \textbf{State 2.1} is a working cycle. First, a \textbf{Program Generator} (not shown in Figure \ref{fig:walkthrough_tree_search}, realized by a UnifiedQA-T5-base) takes \textbf{Episodic Memory (State 1)} as the input and generates a short program, \textbf{Program 1}. Then the \textbf{Program Interpreter} (not shown in Figure \ref{fig:walkthrough_tree_search}) parses and executes \textbf{Program 1}, which results in updating the \textbf{Episodic Memory} and leads to \textbf{Episodic Memory (State 2.1)}. The dashed line means some working cycles between the two states are omitted in the figure.

\paragraph{Initialization:} The \textbf{External Memory} and \textbf{Episodic Memory} are initialized before the first working cycle. The external memory is initialized with the context of the problem. For the tree search problem, the context contains the facts and rules (as shown in Table \ref{tab:dataset_examples}). The content in the external memory is not changed during the working cycles of solving a tree search problem, and the content in the external memory does not contribute to the generation of the program. But the content in the external memory can be accessed by the \texttt{get$\_$statement()} operator, and might be needed in the execution of the program (e.g., \textbf{Program 3.1}). 

The \textbf{Episodic Memory} is initialized with the question (``Did Matthew Smith have 8 bananas?'') and some simple textual description of the task (``This is a tree search task.''), leading to \textbf{Episodic Memory (State 1)}.

\paragraph{Episodic Memory State 1 $\rightarrow$ 2.1: } 
The program generator takes the \textbf{Episodic Memory (State 1)} as the input and generates \textbf{Program 1}. The input to the program generator (i.e., a trained UnifiedQA-T5-base) is \texttt{generate$\_$program: [text in episodic memory]}. Program 1 contains two lines: first, a sub-question ``Which chunk can prove Matthew Smith had 8 bananas?'' is generated and assigned to a local variable \texttt{$\#$0}, then the \texttt{new$\_$mem($\#$0)} operator is generated to start a new recursion process to address the sub-question stored in $\#$0. \textbf{Program 1} is parsed and executed by the interpreter, leading to \textbf{Episodic Memory (State 2.1). }

\paragraph{Episodic Memory State 2.1 $\rightarrow$ 3.1: } The program generator takes \textbf{Episodic Memory (State 2.1)} as the input and generates \textbf{Program 2.1}. \textbf{Program 2.1} loops over all chunks. For each chunk, a sub-question ``Can this chunk prove Matthew Smith have 8 bananas?''. The sub-question is further rewritten by the \texttt{rewrite} module as ``Can chunk $k$ be used to prove Matthew Smith had 8 bananas'' is generated. The \texttt{rewrite} module is also a text-to-text neural network, realized by a UnifiedQA-T5-base model. The input to the module is \texttt{rewrite: [text piece 1] [text piece 2]}. For each generated sub-question, a new recursion process is started by the \texttt{new$\_$mem()} operator.

Note that we do not restrict the types and depths of recursion. The program generator is allowed to generate the program \texttt{new$\_$mem()} to start another recursion, as long as it helps with solving the problem. The ability to generate multiple depths and types of recursion calls is one of the major difference between EVR+ and other problem decomposition methods. A brief discussion is given in Section \ref{sec:discussion}.

\paragraph{Episodic Memory State 3.1 $\rightarrow$ 4.1} For each chunk, a \texttt{qa} module (a text-to-text neural model, realized by a UnifiedQA-T5-base) is called to answer the question ``Can chunk 0 be used to prove Matthew Smith had 8 bananas?''. To answer the question, the \texttt{qa} module also gets the context of the problem from the \textbf{External Memory} with the operator \texttt{get$\_$chunk('chunk$\_$0')}. Therefore the input to the \texttt{qa} module is \texttt{qa: [sentences in chunk 0] [question]}. Again the program generator first generates \textbf{Program 3.1}, then the program is executed. After execution, the result is added to the episodic memory, leading to \textbf{Episodic Memory (State 4.1)}. 

\paragraph{Episodic Memory State 4.1 $\rightarrow$ 2.2} The program generator generates the operator \texttt{return(episodic$\_$buffer$\_$1)} to terminate the recursion process and return the results stored in \texttt{episodic$\_$buffer$\_$1} to the upper level. This leads to \textbf{Episodic Memory (State 2.2).}

\paragraph{Episodic Memory State 2.2 $\rightarrow$ Result} More working cycles happen until the result is obtained. In each working cycle, the program generator generates a short piece of program based on the episodic memory, and the program interpreter executes the program, which updates the episodic memory. For a more detailed execution flow of the tree search problem, please check Figure \ref{fig:app_tree_search_design_1} and \ref{fig:app_tree_search_design_2} in Appendix \ref{sec:app_execution_flows}.

\subsection{Training EVR+}
\label{sec:framework_training}
The example in Figure \ref{fig:walkthrough_tree_search} involves three neural components: \texttt{program$\_$generator}, \texttt{qa} module and \texttt{rewrite} module. All these three modules are implemented using the same UnifiedQA-T5-base model. For training, we use hand-crafted rules to generate the training data for each module. For example, the program generation from \textbf{Episodic Memory (State 1)} to \textbf{Program 1} is a text-to-text prediction problem. The input is the episodic memory and the target output is the program. We call this \textit{pattern} of training data \textit{generate$\_$program-1}. Similarly, the generation from \textbf{Episodic Memory (State 2.1)} to \textbf{Program 2.1} is \textit{generate$\_$program-2}, the training data pattern of the \texttt{qa} module is \textit{qa-1}, and the training data pattern of the \texttt{rewrite} module is \textit{rewrite-1}. In total, we generated 12 patterns of training data for the tree search task. We mix all of the patterns of training data and train only one UnifiedQA-T5-base model to fulfill all neural module's functionalities. Different patterns of training data have different prefixes: e.g., the \texttt{qa} data has a \texttt{qa} prefix. More detailed descriptions and examples of the training data can be found in Appendix \ref{sec:app_evr_training_data}.

Note that the ``State'' in Figure \ref{fig:walkthrough_tree_search} does not mean everything is deterministic: since \textbf{State 2.1} is the result of the execution of \textbf{Program 1} and \textbf{Program 1} is the generated (i.e., predicted) text by the UnifiedQA-T5-base model, \textbf{State 2.1} can also be viewed as a prediction from \textbf{2.2}.

\section{Experiments}
After training EVR+ with the protocol introduced in Section \ref{sec:framework_training}, we evaluate EVR+ on all of the 5 tasks in the SynthCompR dataset.
\paragraph{Baseline} We use an end-to-end trained UnifiedQA-T5-large as the baseline for all tasks. The UnifiedQA-T5-large model uses \texttt{[question] \textbackslash n [context']} as the input, and is trained to directly predict the answer as plain text. Here \texttt{[context']} is built from the \textit{context} of the problem (such as in Table \ref{tab:dataset_examples}), but slightly modified to ensure it has the same information in the input as EVR+.\footnote{Please check Appendix \ref{sec:app_unifiedqa_input_format} for details.}

We design the main experiments in this section along the following research questions:
(1) Does EVR+ achieve better compositional generalization performance compared with the end-to-end trained UnifiedQA model? (2) Does EVR+ require less data when adapting from simpler tasks to more complex tasks? (3) What is the potential of EVR+ to be used for other Large Language Models (LLMs) with in-context learning?

\subsection{Reasoning Depth Generalization}

\begin{figure*}[!ht]
    \centering
    \includegraphics[width=0.98\textwidth]{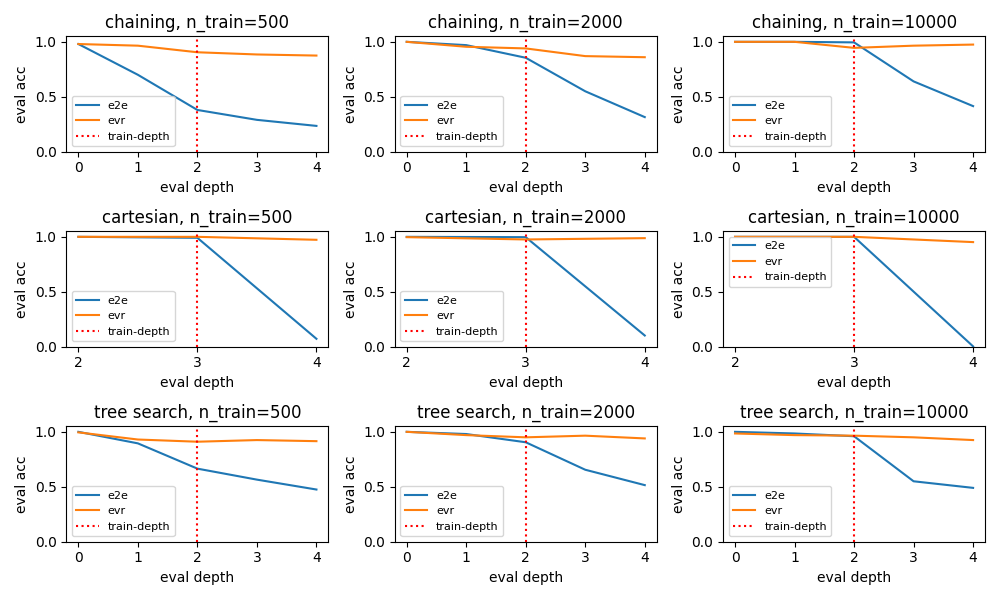}
    \caption{The performance of UnifiedQA-T5-large (e2e) and EVR+ (evr, with a UnifiedQA-T5-base backbone) on the chaining, Cartesian and tree search tasks. The e2e model is trained on depth up to 2 data for chaining and tree search (i.e., data depth 0, 1, 2), and trained on depth up to 3 data for Cartesian product (i.e., depth 2, 3).}
    \label{fig:exp1}
\end{figure*}

We train a UnifiedQA-T5-large model end-to-end and train an EVR+ model (with a backbone model of UnifiedQA-T5-base\footnote{EVR+ needs to run a language model many times for one problem. A T5-Base model largely reduces the inference time.}) on the chaining, Cartesian and tree search datasets. 
Importantly, both models are trained on the data with depth up to 2 for chaining and tree search, and on data with depth up to 3 for Cartesian product.\footnote{The Cartesian task with depth 1 will be similar to: ``John Smith had 2 apples. List the items that each person had.'', and the answer will be the copy of the first sentence. Such an example is not very meaningful, so we set the minimum depth of the Cartesian product to be 2. We also want to include the data with at least two depths in the training set, so that it is reasonable to expect the model to learn to generalize. Eventually we set the training depth of the Cartesian product task to be 2 and 3.} For chaining and tree search, the data with depth of 3 and 4 are out-of-domain data (OOD); for Cartesian product, the data with the depth 4 is OOD data. This way we explicitly test the model's compositional generalization performance on OOD data. We also experimented with using different number of training examples (i.e., $\{$500, 2000, 10000$\}$). Figure \ref{fig:exp1} shows the evaluation accuracy on the test set with different depths and different number of training examples. 
We draw several observations from this experiment:

\textbf{The end-to-end model is more hungry for training examples.} On the chaining and tree search tasks, 500 training examples are not enough for the end-to-end model to learn to solve the problems even for the in-domain data. As the number of training examples increase to 10000, the end-to-end model is able to learn the in-domain data well. In contrast, EVR+ is less data hungry, as it can learn the data well with only 500 training examples.

\textbf{EVR+ generalizes much better than the end-to-end model when tested on out-of-domain data.} On all three tasks, the out-of-domain performance for the end-to-end model drops significantly, even with abundant training examples, but the EVR+ model is able to achieve almost the same performance as on the in-domain data, with as few as 500 training examples. Note that in all experiments the EVR+ uses a smaller language model than the end-to-end trained model.

\subsection{Task Composition}

\begin{figure}[!ht]
    \centering
    \includegraphics[width=0.48\textwidth]{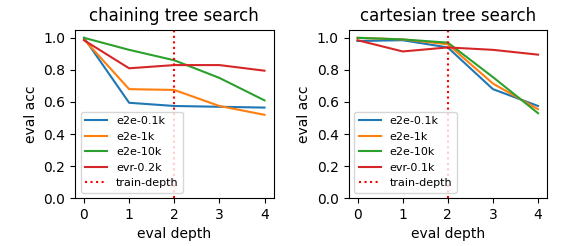}
    \caption{The performance of UnifiedQA-T5-large (e2e) and EVR+ (evr, with a UnifiedQA-T5-base backbone) on the chaining-tree-search and Cartesian-tree-search tasks. For both tasks the models are trained on the data with the depth up to 2, so the data with depth 3 and 4 are OOD data.}
    \label{fig:exp2}
\end{figure}

We further evaluate how easy it is for the model to generalize from easy tasks to complex tasks using the chaining tree search task and Cartesian tree search task. For the chaining tree search task, we first train both models on the mixed chaining examples and tree search examples (with 10,000 examples for each task). Then we fine-tune the end-to-end model on the chaining tree search task using $\{$100, 1000, 10000$\}$ examples, and fine-tune the EVR+ model on 200 chaining tree search tasks. 
Because 100 examples are not sufficient to train EVR+ to learn all  patterns, we skip this setting, and, instead, evaluate EVR+ fine-tuned on 200 examples. 
A similar protocol is applied to the Cartesian tree search task, where we first train the models on the mixture of Cartesian product and tree search examples, then fine-tune the end-to-end model on the Cartesian tree search task with $\{$100, 1000, 10000$\}$ examples, and fine-tuned the EVR+ model on 100 Cartesian tree search examples. 

\textbf{EVR+ needs little data for transfer learning of the chaining tree search task.} Figure \ref{fig:exp2} (left) shows the performance of UnifiedQA-T5-Large against EVR+ (with a UnifiedQA-T5-Base backbone). EVR+ fine-tuned on 200 examples has a equally good or better performance than the end-to-end model fine-tuned on 1,000 examples on all depth. Even when the end-to-end model is fine-tuned on 10,000 examples, the EVR+ fine-tuned on 200 examples still has better performance on depth 3 and 4. 

\textbf{EVR+ with 100 fine-tuning examples achieves better OOD performance than the end-to-end method on the Cartesian tree search task.} Figure \ref{fig:exp2} (right) shows the performance of the two models trained on the Cartesian tree search task. Results show that the end-to-end model can transfer pretty well from the Cartesian product and tree search task to the Cartesian tree search task. The end-to-end model is able to achieve nearly 1.0 test accuray on the in-domain test set (depth 0, 1, 2). The EVR+ model can mostly learn the task with only 100 fine-tuning examples, but the in-domain performance is not as high as the end-to-end model. However, the out-of-domain performance of the EVR+ model is still much better than the end-to-end model, even if the end-to-end model uses 10,000 fine-tuning data.

\subsection{EVR+'s Compatibility with LLMs}
\label{sec:llm_compatibility}

\begin{table}[!ht]
    \centering
    \begin{tabular}{P{5.5cm} | p{1.5cm}} \hline
        Patterns & Test Acc. \\ \hline \hline
        qa-1, qa-2, rewrite-1, clear$\_$mem-1, generate$\_$program-$\{$2,3,5,6,9$\}$ & 1.0 \\ \hline
        clear$\_$mem-2, generate$\_$program-$\{$4, 7, 8$\}$ & >0.5 \\ \hline
        generate$\_$program-$\{$1, 10, 11$\}$ & <0.2 \\ \hline
    \end{tabular}
    \caption{Few-shot prompted test accuracy for each pattern. }
    \label{tab:exp3}
\end{table}

\begin{figure*}
    \centering
    \begin{subfigure}[b]{0.49\textwidth}
        \centering
         \includegraphics[width=\textwidth]{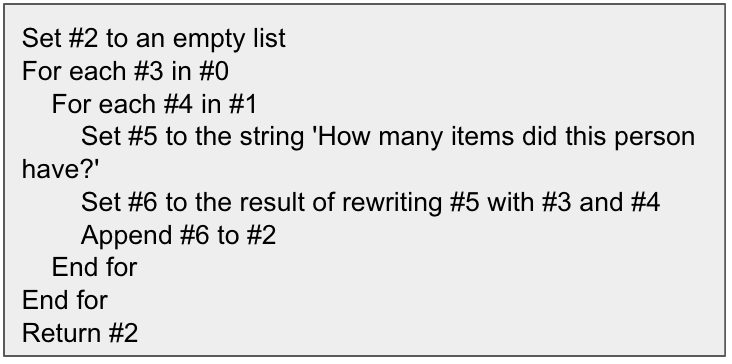}
         \caption{Program Generated by GPT-3.}
         \label{fig:carte_few_shot_gen}
     \end{subfigure}
     \hfill
     \begin{subfigure}[b]{0.49\textwidth}
         \centering
         \includegraphics[width=\textwidth]{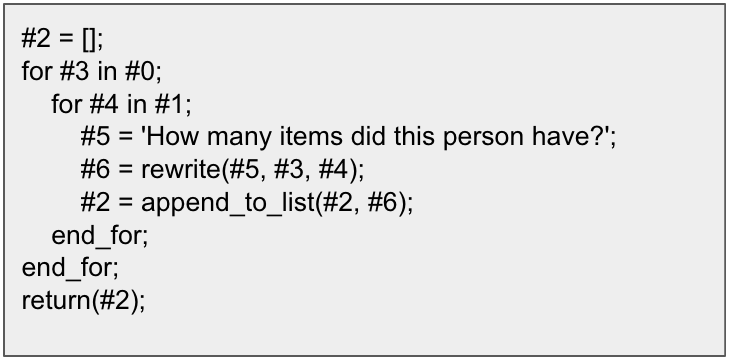}
         \caption{Target Program.}
         \label{fig:carte_few_shot_tar}
     \end{subfigure}
    \caption{The program generated by GPT-3 and the target program. All lines of the program generated by GPT-3 are aligned with the target program, but with incorrect grammar.}
    \label{fig:carte_few_shot}
\end{figure*}

The results in Figure \ref{fig:exp1} and Figure \ref{fig:exp2} are obtained by fine-tuning the UnifiedQA model \cite{khashabi2020unifiedqa}  on each pattern of EVR+ training data. In this section we discuss the possibility of replacing the fine-tuned UnifiedQA model with a few-shot prompted large language model, such as GPT-3 \cite{brown2020language}. We use the Cartesian product task as the evaluation task in this section.

First, we assume the EVR+ training data are available for 4 Cartesian product examples (i.e., 4 annotated examples): 2 with depth 2 and 2 with depth 3. Each Cartesian product example will result in 16 patterns of training data. This results in 172 training examples in total. 

Then for each pattern of data, we sample 10 examples\footnote{We did not use all test examples due to the budget limit.} from the test set with depth 4. Note that in this case the test data are OOD because the training data are from depth 2 and 3. For each sample test example, we use a sentence-T5 encoder \cite{ni2021sentence} to retrieve 4 the most similar training examples to the test example and form the prompt (the retrieval of few-shot examples is similar to the method introduced in \cite{liu2021makes}). Note that the sentence-T5 encoder might retrieve the wrong pattern for the test example. For example, a \textit{generate$\_$program-1} test example might retrieve a \textit{generate$\_$program-2} few-shot example. This is to emulate the situation when EVR+ is used at inference. That is, at inference time, the text in the episodic buffer will not indicate what patterns of data EVR+ should retrieve as the few-shot example. Instead, EVR+ needs to retrieve the most similar examples from all patterns.

Table \ref{tab:exp3} shows the test accuracy by the pattern. More than half of the patterns can achieve 1.0 test accuracy using a few-shot prompted GPT-3 model. However, some patterns are very hard to be correctly generated, such as \textit{generate$\_$program-$\{$1, 10, 11$\}$}. One major mistake for the model we notice is that GPT-3 tends to generate logically reasonable program without the correct grammar. Figure \ref{fig:carte_few_shot} shows the generated program by GPT-3 for pattern \textit{generate$\_$program-11} and the target program. This is somehow expected because the program we used for our reasoning framework is newly defined. It is hard for the model to generate the program perfectly with only a few exemplars in the input prompt. Overall, our framework has the potential to be combined with few-shot prompted language models, but some modifications are needed. We leave this direction as future work. 

\section{Discussion}
\label{sec:discussion}
\subsection{Relation with Other Decomposition Methods}
\label{sec:relation_with_other_methods}
%



Our reasoning framework differs from previous decomposition frameworks because it supports the generation of actions both sequentially and recursively. The most similar reasoning framework to ours is \cite{khot2022decomposed}, where both sequential and recursive decompositions are supported. Our method differs from their methods in three ways. First, the expressions supported by our framework is much more diverse than in \cite{khot2022decomposed}. For example, the program \textbf{from State 11 to the Final Result} in Figure \ref{fig:app_Cartesian_design} in Appendix \ref{sec:app_execution_flows} contains a nested loop, and some additional operations that happen in the body of the loop (e.g., \texttt{rewrite}, \texttt{append$\_$to$\_$list}). Our interpreter is able to interpret and execute such programs, whereas such an ability is not demonstrated in the other directions above.

Second, the recursion operator \texttt{new$\_$mem()} in our system is a very general description of any recursion process. The \texttt{new$\_$mem()} operator can be used to start different recursion processes with different purposes (for examples, please refer to the various usages of \texttt{new$\_$mem()} in Section \ref{sec:walkthrough} and Appendix \ref{sec:app_execution_flows}). In contrast, the recursion operations defined by \cite{khot2022decomposed} is a task specific one, as it is only used to handle the list reversal task. 

Finally, our framework defines the operators that can directly manipulate the memory such as the external memory and the episodic memory. One example is in Figure \ref{fig:walkthrough_tree_search} \textbf{Program 3.1}, where the \texttt{get$\_$chunk()} operation is used to read the content from the external memory and load the content into local variables (between state 2 and state 3), and the \texttt{clear$\_$mem()} operation is used to eliminate the redundant information in the episodic memory (from state 10 to state 11).

\subsection{Relation with Python}
\label{sec:relation_with_python}
The generated program in our system is fairly similar to python. During the developing of our program and interpreter, we gradually added more and more supported expressions, and eventually we find composing the program in a python-like style results in very natural flows. In the future we plan to modify our interpreter and use python as the generated program. This will also hopefully reduce the grammatical issue of the few-shot generated programs mentioned in Section \ref{sec:llm_compatibility}. 

\section{Conclusion}
In this paper we introduced Explainable Verbal Reasoner Plus (EVR+), a reasoning framework featuring (1) the ability to generate and execute diverse symbolic operations and (2) the ability to solve diverse types of compositional reasoning problems through problem decomposition. To test EVR+ performance, we proposed SynthCompR, a synthetic dataset containing 5 different tasks that require compositional reasoning. Results show that EVR+ with a smaller language model can generally outperform a larger end-to-end trained model. Finally, we analyzed the relation between EVR+ and other decomposition-based methods, as well as some limitations and future directions. 
All in all, our work indicates that modeling compositionality formally is feasible and leads to better generalization performance.

\section{Limitations}
In this paper the EVR+ framework is only tested on synthetic tasks. On real-world datasets the cascading errors might have a huge negative impact of the model's performance (as previous study shows, even composing two sentences can sometimes be hard for the language model \cite{liang2020transformers}). In addition, the inference time of EVR+ is much higher than the regular end-to-end trained model with the same size, because EVR+ needs to run the language model multiple times when handling one problem. For example, a chaining problem with depth 4 requires to run the UnifiedQA-T5-base model 26 to 27 times. 

\bibliography{anthology,custom}
\bibliographystyle{acl_natbib}

\newpage
\appendix
\onecolumn

\section{Dataset Construction}
\label{sec:app_dataset_construction}
For all the 5 tasks, we use rules to compose the formal representations and use a very simple template to convert the formal representations to natural language expressions. For example, the formal representation \texttt{(Matthew Smith, 8, bananas)} is converted to the natural language expression \texttt{Matthew Smith had 8 bananas}.

\subsection{Names, Items and Quantities}
Below is what we used to get the names, items and quantities that appear in the dataset.
\begin{itemize}
    \item \textbf{Name}: We get the top 1000 most popular first and last names from \url{https://namecensus.com/} and combine each first name and last name. This result in 1 million unique full names. 
    \item \textbf{Item}: We include 12 different items, such as apples, pears, toy bears, etc., in our dataset.
    \item \textbf{Quantity}: We restrict the quantities to be from 0 to 20, only integers. 
\end{itemize}

\subsection{Construction Logic for Each Task}
Below are the high level descriptions of the method to construct the dataset. The detailed implementation can be found at: \\
\url{https://github.com/zhengzhongliang/ExplainableVerbalReasonerPlus}.

\begin{itemize}
    \item \textbf{Chaining}: For each chaining problem, we build 3 to 5 chains. For each chain, we first randomly sample a triple \texttt{(main$\_$name, quantity, item)} as the initial statement, which will be translate to \texttt{main$\_$name had quantity items}. Then we sample $d$ other statements where each statement is \texttt{(other$\_$name, delta$\_$quant, item)} and will be later translate to \texttt{other$\_$name gave main$\_$name delta$\_$quant items}. Here $d$ is the depth of the problem.

    \item \textbf{Cartesian Product}: For each Cartesian product problem, we sample $d$ names, $d$ quantities and $d$ items, where $d$ is the depth of the problem. Then we translate the sampled representations to natural language, such as \texttt{Each of person1, person2 and person3 had X apples, Y pears and Z toy bears}.

    \item \textbf{Tree Search}: For each tree search problem, we first sample $k_1$ grounded statements and $k_2$ ungrounded statements. Both the grounded and ungrounded statements are triples with the form \texttt{(name, quantity, item)}. After the sampling of the initial statements, we construct the example in a forward chaining manner: for each reasoning step, we generate $k_3$ rules, where each rule has the format \texttt{preconditions $\rightarrow$ conclusion}. The preconditions have 1 or 2 statements (each statement is triple), and the conclusion is a single triple. Among these $k_3$ rules, at least 1 rule uses the grounded statements as the preconditions and therefore leading to a grounded conclusion (which is a new grounded statement). Meanwhile at least 1 rule uses the ungrounded statements as the precondition and therefore leading to an ungrounded conclusion (a new ungrounded statement). Such a reasoning step is repeated for $d$ times, where $d$ is the depth of the problem, to generate more grounded and ungrounded statements. After we get the grounded and ungrounded steps generated from the $d^{th}$ step, we select what statement should be used to generate the question: with a 0.5 probability the question is about the grounded statement of the $d^{th}$ step (i.e., leading to the final answer \texttt{Yes}), and with a 0.5 probability the question is about the ungrounded statements of the $d^{th}$ step (i.e., leading to the final answer \texttt{No}). 

    \item \textbf{Chaining Tree Search}: The generation process is similar to the tree search task. But in the beginning, we use the statements that are entailed from the chaining task as the initial statements.

    \item \textbf{Cartesian Tree Search}: The generation process is similar to the tree search task. But in the beginning, we use the statements that are entialed from Cartesian task as the initial statements. 
\end{itemize}

\newpage
\section{List of Supported Operations}
\label{sec:appendix_operations}
\begin{table*}[ht!]
    \centering
    \begin{tabular}{P{1.5cm} | p{4cm} | p{9.5cm}} \hline
        Category & Example & Comment \\ \hline \hline
        
        Constant & 1, 2, 3, ... &  Integer constants \\ \cline{2-3}
                 & 'text 1.', 'text 2.' & String constants, always wrapped by single quotations. \\ \cline{2-3}
                 & True, False  & Boolean constants \\ \hline

        List & [1,2,3], ['text 1.', 'text 2.']; & The list can have elements of constants, enclosed by brackets and separated by commas. \\ \hline
                 
        Variable & $\#$0=2; $\#$1='some text.';  & The local variable has the form $\#$0, $\#$1, etc.. The expressions assign some values to the local variable $\#0$ and $\#1$.  \\ \hline
        
        Control Flow & \makecell[l]{for $\#$1 in $\#$0; \\ \; $\#$2 = $\#$1; \\ end$\_$for; }  & Loop over all elements in the list $\#$1. A for loop must end with end$\_$for. \\ \cline{2-3}

         & \makecell[l]{while [condition]; \\ \; LOOP$\_$BODY \\ end$\_$while;}  & Loop while the condition is true.  \\
        \cline{2-3}

        & \makecell[l]{if [condition]; \\ \; COND$\_$BODY1 \\ else; \\ \; COND$\_$BODY2 \\ end$\_$if;}  & Execute body 1 or 2 based on the condition.  \\
        \hline
        
        Memory Operation & new$\_$mem($\#$1, $\#$2);  & Starts a new recursion function. The value of variable $\#$1 and $\#$2 will be used to initialize the episodic memory of the new recursion process. \\ \cline{2-3}

        & return($\#$1);  & Terminate the current level of recursion and add the value of variable $\#$1 to the parent level's episodic buffer. \\ \cline{2-3} 

         & add$\_$to$\_$episodic($\#$1); & Get the content in the variable $\#$1 and append it to the episodic memory \\ 
        \cline{2-3}

        & clear$\_$mem(); & Eliminate the redundant information in the episodic memory. \\ 
        \cline{2-3}

        & check$\_$next$\_$atatement() & Loop over the statements of the one chunk in the external memory. Return true if this is not the last statement. \\ \cline{2-3}

        & get$\_$next$\_$statement$\_$num() & Get the next statement's name, e.g., 'statement$\_$0'. \\ \cline{2-3}

        & get$\_$statement('statement$\_$0') & Get the content of 'statement$\_$0' \\ \cline{2-3}

        & check$\_$next$\_$chunk() & Loop over the chunks of the external memory. Return true if this is not the last chunk. \\ \cline{2-3}

        & get$\_$next$\_$chunk$\_$num() & Get the next chunk's name, e.g., 'chunk$\_$0'. \\ \cline{2-3}

        & get$\_$chunk('chunk$\_$0') & Get the content of 'chunk$\_$0' \\ \cline{2-3}

        & list$\_$chunks('chunk$\_$0', 'chunk$\_$3')  & List the chunks names from 'chunk$\_$0' to 'chunk$\_$3' (both included). \\ \cline{2-3}

        & update$\_$chunk('chunk$\_$0', $\#$1)  & Update 'chunk$\_$0' with the value stored in variable $\#$1. \\ \cline{2-3}

        & clean$\_$chunks()  & If any chunk has more than 3 statements, split that chunk to multiple chunks, with each new chunk having at most 3 statements. \\ \cline{2-3}

        & del($\#$1) & Delete the local variable $\#$1 \\ \hline
        
    \end{tabular}
    \caption{A List of Supported Expressions (Part 1)}
    \label{tab:list_of_expressions_1}
\end{table*}

\newpage
\begin{table*}[ht!]
    \centering
    \begin{tabular}{P{1.5cm} | p{4cm} | p{9.5cm}} \hline
        Category & Example & Comment \\ \hline \hline
        Utility & append$\_$to$\_$list($\#$1, $\#$2) & Append the content of variable $\#$2 to the list $\#$1. \\ \hline

        External Tools & qa('John had 2 apples. John gave Mike 1 apple.', 'How many apples did John have?'); & An question answering module. The first argument is the context and the second argument is the question. The QA module returns the answer of the question based on the context. \\ \cline{2-3}

        & rewrite('How many apples did John have?', '1'); & An question answering module. The first argument is the context and the second argument is the question. The QA module returns the answer of the question based on the context. \\ \cline{2-3}

        & subqs() & Generate some sub-goals based on the content in the episodic memory or by some other content. \\ \hline
    \end{tabular}
    \caption{A List of Supported Expressions (Part 2)}
    \label{tab:list_of_expressions_2}
\end{table*}

\newpage
\section{Detailed Execution Flow of Each Task}
\label{sec:app_execution_flows}
\subsection{Chaining}
\begin{figure*}[ht!]
\centering
    \includegraphics[width=0.95\textwidth]{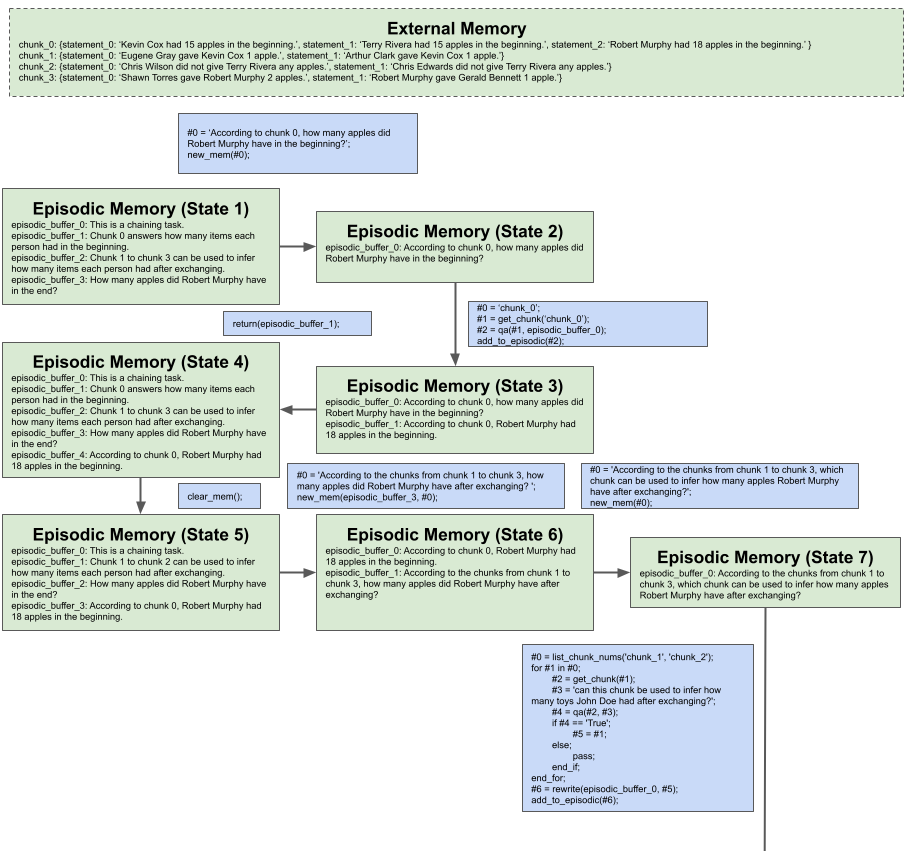}
    \caption{Execution Flow of the Chaining Task (Part 1).}
    \label{fig:app_chaining_design_1}
\end{figure*}

\newpage
\begin{figure*}[ht!]
\centering
    \includegraphics[width=0.95\textwidth]{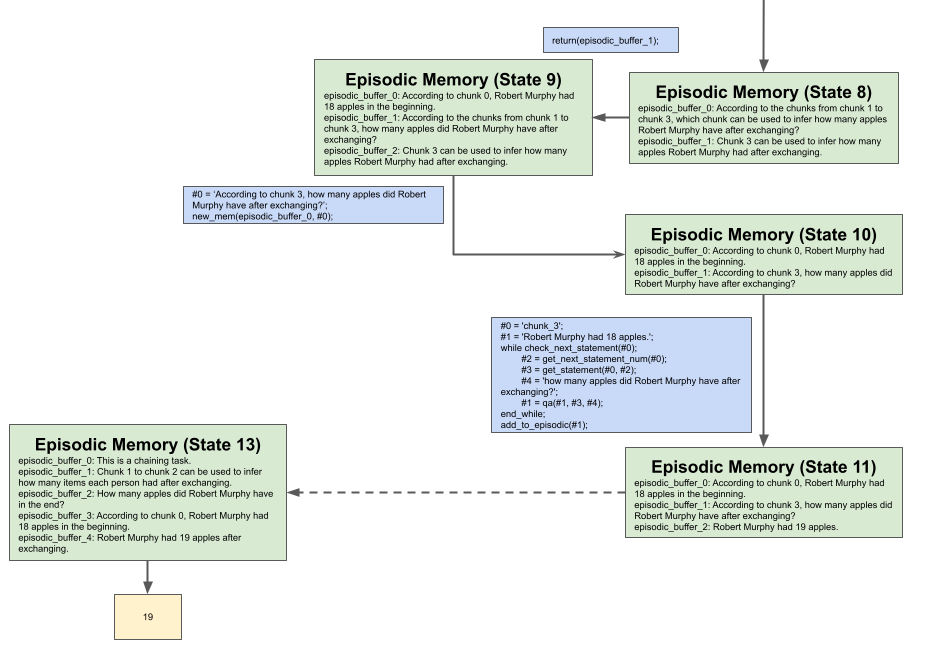}
    \caption{Execution Flow of the Chaining Task (Part 2).}
    \label{fig:app_chaining_design_2}
\end{figure*}

\newpage
\subsection{Carteisan Product}
\begin{figure*}[ht!]
\centering
    \includegraphics[width=0.95\textwidth]{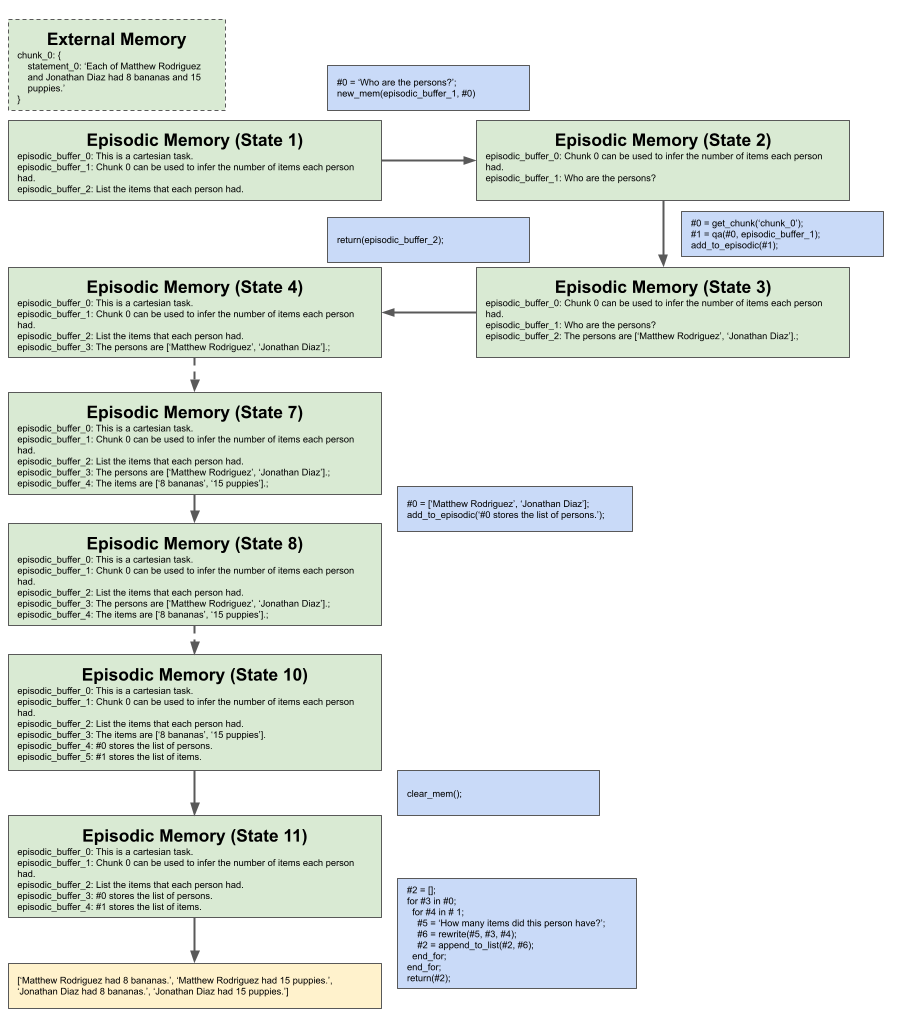}
    \caption{Execution Flow of the Cartesian Product Task.}
    \label{fig:app_Cartesian_design}
\end{figure*}

\newpage
\subsection{Tree Search}
\begin{figure*}[ht!]
\centering
    \includegraphics[width=0.95\textwidth]{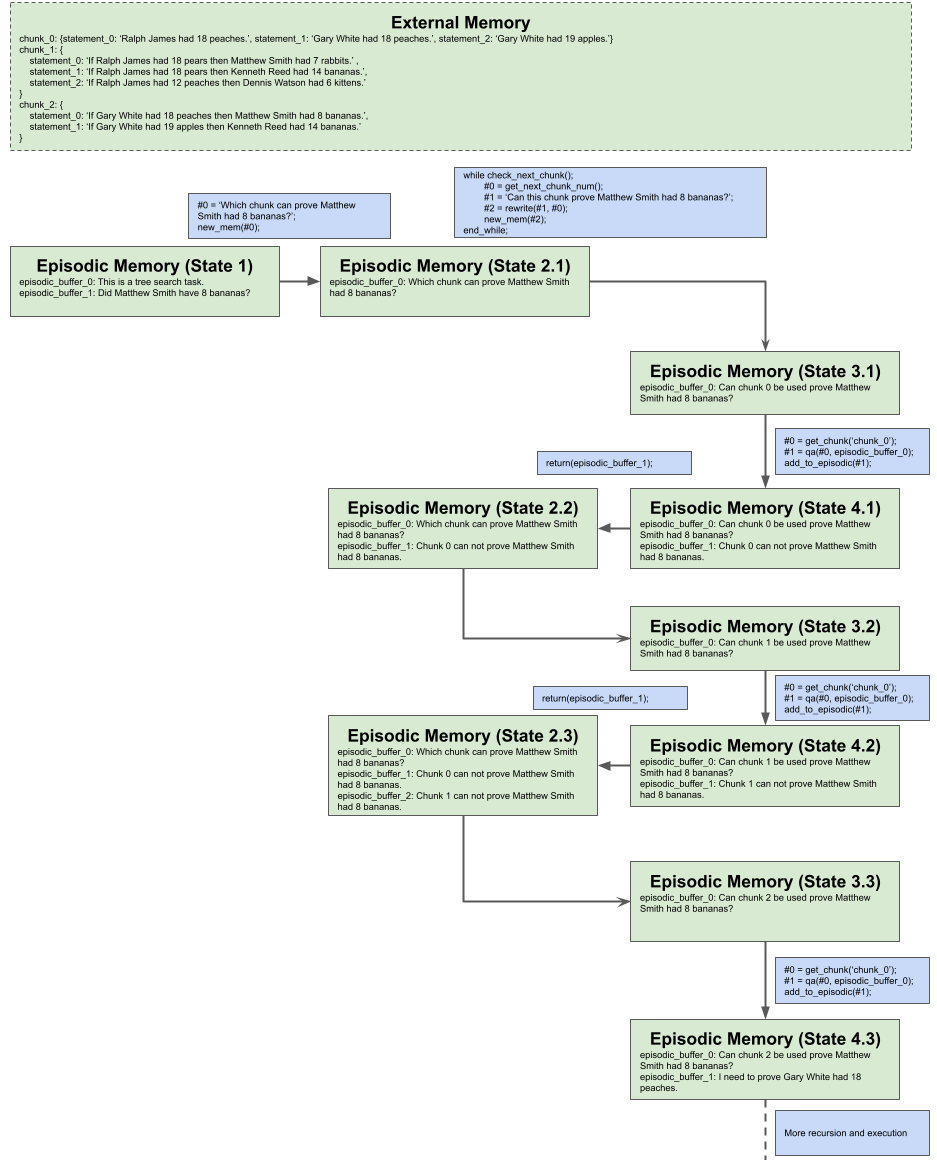}
    \caption{Execution Flow of the Tree Search Task (Part 1).}
    \label{fig:app_tree_search_design_1}
\end{figure*}

\newpage
\begin{figure*}[ht!]
\centering
    \includegraphics[width=0.95\textwidth]{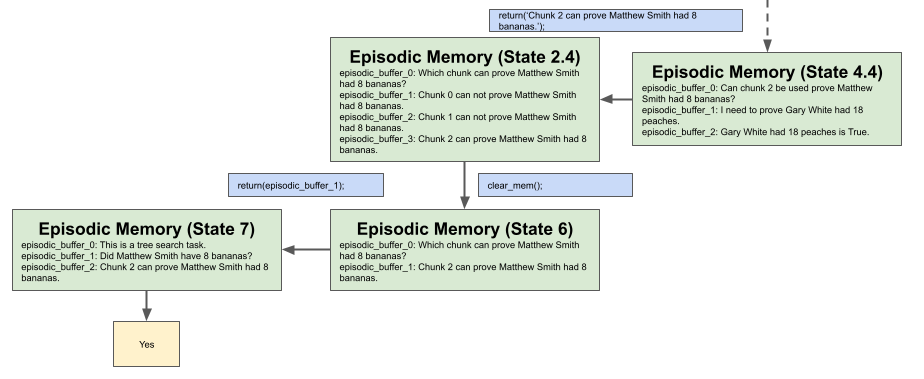}
    \caption{Execution Flow of the Tree Search Task (Part 2).}
    \label{fig:app_tree_search_design_2}
\end{figure*}

\newpage
\subsection{Chaining Tree Search}
\begin{figure*}[ht!]
\centering
    \includegraphics[width=0.95\textwidth]{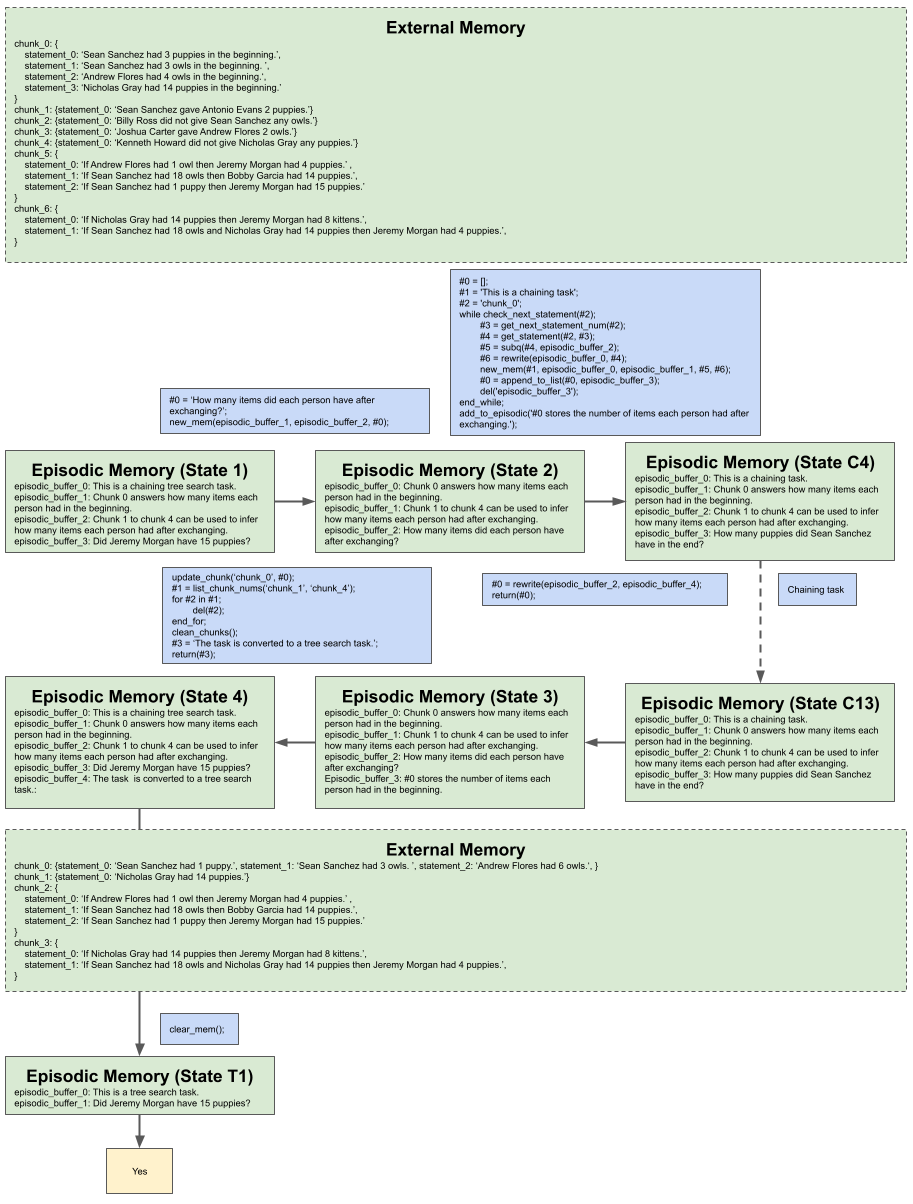}
    \caption{Execution Flow of the Chaining Tree Search Task.}
    \label{fig:app_chaining_tree_search_design}
\end{figure*}

\newpage
\subsection{Cartesian Tree Search}
\begin{figure*}[ht!]
\centering
    \includegraphics[width=0.95\textwidth]{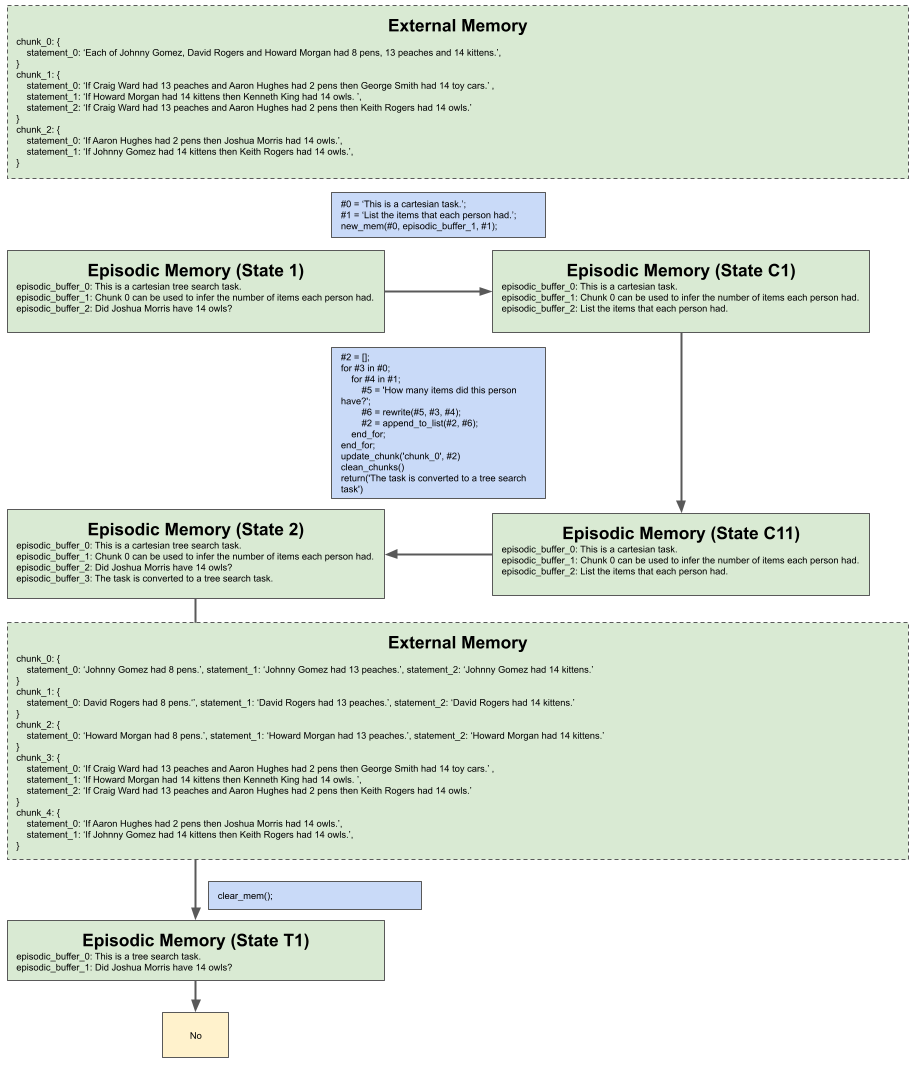}
    \caption{Execution Flow of the Cartesian Tree Search Task.}
    \label{fig:app_Cartesian_tree_search_design}
\end{figure*}

\newpage
\section{Generating Training Data for EVR+}
\label{sec:app_evr_training_data}
As stated in Section \ref{sec:framework_training}, we train the neural modules (all realized by a single UnifiedQA-T5-base) by patterns. For example, in Figure \ref{fig:walkthrough_tree_search}, the mapping from \textbf{Episodic Memory (State 1)} to \textbf{Program 1} is pattern \textit{generate$\_$program-1}; the mapping from \textbf{Episodic Memory (State 2.1)} to \textbf{Program 2.1} is \textit{generate$\_$program-2}. There are 12 patterns of data in total.

We generate all patterns of training examples from each tree search problem with hand-crafted rules. Table 
\ref{tab:app_tree_search_training_data_gen} shows the hand-crafted template to generate the training data of 3 patterns out of 12 patterns.

\begin{table}[!ht]
    \centering
    \begin{tabular}{ p{3.5cm} | p{1cm} | p{10cm} } \hline
        pattern & field & content \\ \hline \hline
        generate$\_$program-1 & input & \makecell[l]{generate$\_$program: \\ episodic$\_$buffer$\_$0: This is a tree search task. \\ episodic$\_$buffer$\_$1: [tree search question]} \\ \cline{2-3}
            & target & \makecell[l]{$\#$0 = 'Which chunk can prove [tree search question]?'; \\ new$\_$mem($\#$0);} \\  \hline

        generate$\_$program-2 & input & \makecell[l]{generate$\_$program: \\ episodic$\_$buffer$\_$0: Which chunk can prove [tree search question] ?} \\ \cline{2-3}
            & target & \makecell[l]{while check$\_$next$\_$chunk(); \\ \; $\#$0 = get$\_$next$\_$chunk$\_$num(); \\ \; $\#$1 = `Can this chunk prove [tree search question] ? `; \\ \; $\#$2 = rewrite($\#$1, $\#$0); \\ \; new$\_$mem($\#$2); \\ end$\_$while;} \\  \hline

        generate$\_$program-3 & input & \makecell[l]{generate$\_$program: \\ episodic$\_$buffer$\_$0: Can chunk [k] be used to prove [tree \\ search question] ?} \\ \cline{2-3}
        & target & \makecell[l]{$\#$0 = get$\_$chunk('chunk$\_$0'); \\ $\#$1 = qa($\#$0, episodic$\_$buffer$\_$0); \\ add$\_$to$\_$episodic($\#$1); } \\  \hline

        qa-1 & input & \makecell[l]{qa: \\ statement$\_$0: [fact or rule] \\ statement$\_$1: [fact or rule]. \\ statement$\_$2: [fact or rule]. \\ Can chunk [k] be used to prove [tree search question]} \\ \cline{2-3}
        & target & \makecell[l]{Chunk [k] can/can not prove [tree search question]} \\  \hline
    \end{tabular}
    \caption{The template to generate the training data generate$\_$program-1, generate$\_$program-2, generate$\_$program-3, qa-1 of the tree search task. The generate$\_$program data have the \texttt{generate$\_$program:} prefix in the input, and the qa data have the \texttt{qa:} prefix in the input.}
    \label{tab:app_tree_search_training_data_gen}
\end{table}

For the examples shown in Table \ref{tab:app_tree_search_training_data_gen}, the content enclosed by \texttt{[} \texttt{]} needs to be replaced with the actual content. Specifically: 

\begin{itemize}
    \item generate$\_$program-1: in \texttt{episodic$\_$buffer$\_$0} of the input, the \texttt{[tree search question]} needs to be replaced with an actual tree search question, e.g., ``Did Matthew Smith have 8 bananas?''. In the output, \texttt{[tree search question]} needs to be replaced with an actual tree search question in the statement form, e.g., ``Matthew Smith had had 8 bananas.''.

    \item generate$\_$program-2: in the input and output, the \texttt{[tree search question]} needs to be replaced with an actual tree search question in the statement form, e.g., ``Matthew Smith had had 8 bananas.''.

    \item generate$\_$program-3: in the input, the \texttt{[tree search question]} needs to be replaced with an actual tree search question in the statement form, e.g., ``Matthew Smith had had 8 bananas.''. \texttt{[k]} needs to be replaces with an actual chunk number, e.g., 0, 1, etc..

    \item qa-1: in the input, the \texttt{[fact or rule]} is the actual tree search context of the tree search problem, e.g., ``Ralph James had 18 peaches.''. In both the input and the output, \texttt{[k]} is the chunk number to deal with, and \texttt{[tree search question]} is the query to prove/disprove. 
\end{itemize}

All patterns of training data for all tasks are generated in a similar manner: use hand crafted rules to construct the templates and fill in the content regarding each problem. For the template of all patterns of training data for all tasks, please go to \url{https://github.com/zhengzhongliang/ExplainableVerbalReasonerPlus}.

\newpage
\section{Input Format of the End-to-end Trained UnifiedQA Baseline}
\label{sec:app_unifiedqa_input_format}
Taking the tree search example, the original \texttt{context} of the problem is: \\

\texttt{Ralph James had 18 peaches. Gary White had 18 peaches. Gary White had 19 apples. If Ralph James had 18 pears then Matthew Smith had 7 rabbits. If Ralph James had 18 pears then Kenneth Reed had 14 bananas. If Ralph James had 12 peaches then Dennis Watson had 6 kittens. If Gary White had 18 peaches then Matthew Smith had 8 bananas. If Gary White had 19 apples then Kenneth Reed had 14 bananas.}

To ensure the same input information is given to EVR+ and the baseline, the \texttt{context'} used as the input of the UnifiedQA-T5-large baseline is:

\texttt{\textbf{This is a tree search task.} \textbf{chunk$\_$0}: Ralph James had 18 peaches. Gary White had 18 peaches. Gary White had 19 apples. \textbf{chunk$\_$1}: If Ralph James had 18 pears then Matthew Smith had 7 rabbits. If Ralph James had 18 pears then Kenneth Reed had 14 bananas. If Ralph James had 12 peaches then Dennis Watson had 6 kittens. \textbf{chunk$\_$2}: If Gary White had 18 peaches then Matthew Smith had 8 bananas. If Gary White had 19 apples then Kenneth Reed had 14 bananas.}

As shown in the text, two changes are made to transform \texttt{context} to \texttt{context'} (shown in \textbf{Bold}). First, the statement `This is a tree search task.` is prepended to the original context, because EVR+ uses this statement to initialize it's episodic memory. Second, The original context is chunked in the same way as the EVR+, and each chunk now has a tag (e.g., \texttt{chunk$\_$0}, \texttt{chunk$\_$1}).

\end{document}